\pdfoutput=1

\documentclass[11pt]{article}
\usepackage{booktabs}
\usepackage{enumitem}
\usepackage{bbding}
\usepackage[final]{acl}
\usepackage{multirow}
\usepackage{times}
\usepackage{latexsym}

\usepackage[T1]{fontenc}

\usepackage[utf8]{inputenc}

\usepackage{microtype}

\usepackage{inconsolata}

\usepackage{graphicx}
\usepackage{listings}
\usepackage{amsmath} 
\usepackage[normalem]{ulem}
\useunder{\uline}{\ul}{}
 \usepackage{booktabs}
\usepackage{colortbl}
 \usepackage[normalem]{ulem}
 \useunder{\uline}{\ul}{}
\title{Mind the Style Gap: Meta-Evaluation of Style and Attribute Transfer Metrics}

\author{Amalie Brogaard Pauli$^1$ \ \ \ \  \ \   Isabelle Augenstein$^2$  \ \ \ \ \ \    Ira Assent$^1$ \\
         $^1$Department of Computer Science, Aarhus University, Denmark  \\ $^2$Department of Computer Science, University of Copenhagen, Denmark  \\  \texttt{\{ampa,ira\}@cs.au.dk, augenstein@di.ku.dk}}

\begin{document}
\maketitle
\begin{abstract} 
Large language models (LLMs) make it easy to rewrite a text in any style -- e.g. to make it more polite, persuasive, or more positive -- but evaluation thereof is not straightforward. A challenge lies in measuring  \emph{content preservation:} that content not attributable to style change is retained. 
This paper presents a large meta-evaluation of metrics for evaluating style and attribute transfer, focusing on content preservation. 
We find that meta-evaluation studies on existing datasets lead to misleading conclusions about the suitability of metrics for content preservation. Widely used metrics show a high correlation with human judgments \emph{despite} being deemed unsuitable for the task -- because they do not abstract from style changes when evaluating content preservation. We show that the overly high correlations with human judgment stem from the nature of the test data. To address this issue, we introduce a new, challenging test set specifically designed for evaluating content preservation metrics for style transfer. We construct the data by creating high variation in the content preservation. Using this dataset, we demonstrate that suitable metrics for content preservation for style transfer indeed are style-aware.
To support efficient evaluation, we propose a new style-aware method that utilises small language models, obtaining a higher alignment with human judgements than prompting a model of a similar size as an autorater.

\end{abstract}

\section{Introduction}
\begin{figure}[t]
    \centering
    \includegraphics[trim=0.75cm 0.75cm 0.75cm 0.75cm,width=0.95\columnwidth]{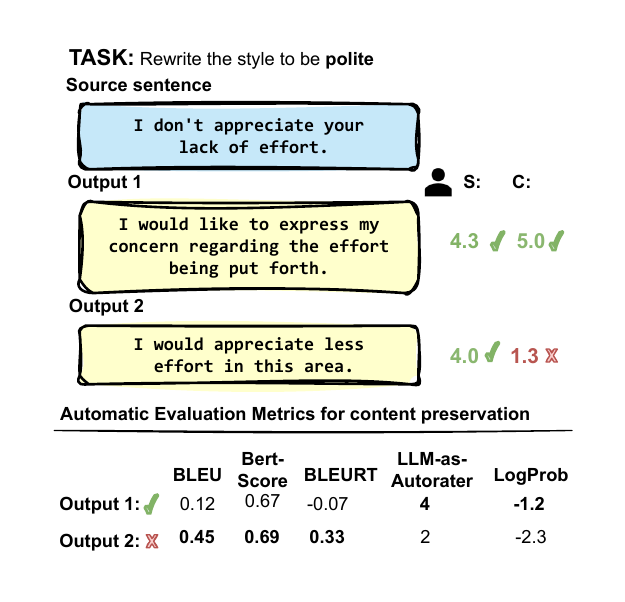}
    \caption{Sample from our new test set with human annotations. S: style strength, C: content preservation. Metrics for content preservation: Bold indicates the highest rated output, and the checkmark indicates successful transfer. }
    \label{fig:sample}
\end{figure}
Large language models allow for rewriting text in a variety of styles or alter any text attribute, without the need for training data. Examples are to make a text more formal, polite, simpler, or to change its sentiment. However, evaluating style and attribute transfer is still challenging as it lacks validation and standardization \cite{ostheimer-etal-2023-call,briakou-etal-2021-review}. 

Traditionally, style and attribute transfer are evaluated via: 1) style strength/shift; 2) fluency and 3) content preservation \cite{jin-etal-2022-deep}. Style transfer papers heavily rely on automatic evaluation metrics \cite{ostheimer-etal-2023-call}. To assess which metrics are most suitable, meta-evaluation efforts measure the correlation of metrics with human judgments.

Measuring content preservation is particularly difficult for style and attribute transfer, as this requires the distinction between content preservation and changes to style or attributes. Still, the status quo is to use metrics not designed for the task of style transfer (most commonly BLEU, \citet{ostheimer-etal-2023-call,jin-etal-2022-deep}), which assess the lexical or semantic similarity of rewrites to the source sentence, not considering style change. Moreover, prior work criticises the use of similarity-based metrics for content preservation \citep{mir-etal-2019-evaluating,lai-etal-2024-style,scialom2021rethinking,logacheva-etal-2022-study}. Intuitively, measuring the similarity between source and output is not suitable for the style transfer task, because the more the style or attribute in the output changes, the more dissimilar the sentences become. Hence, metrics for content preservation should be style-aware \cite{mir-etal-2019-evaluating}. 
    Empirically, however, similarity-based metrics show good correlation with human judgments. In fact, approaches using LLMs only show comparable, and not superior, results to those of semantic similarity metrics \citep{lai2023multidimensional, ostheimer-etal-2024-text}. However, this is even though the LLM-based approaches are a better fit for the style transfer task, as they can be style-aware, as opposed to the similarity-based metrics. 

\begin{figure}[t]
    \centering
    \includegraphics[width=0.95\columnwidth]{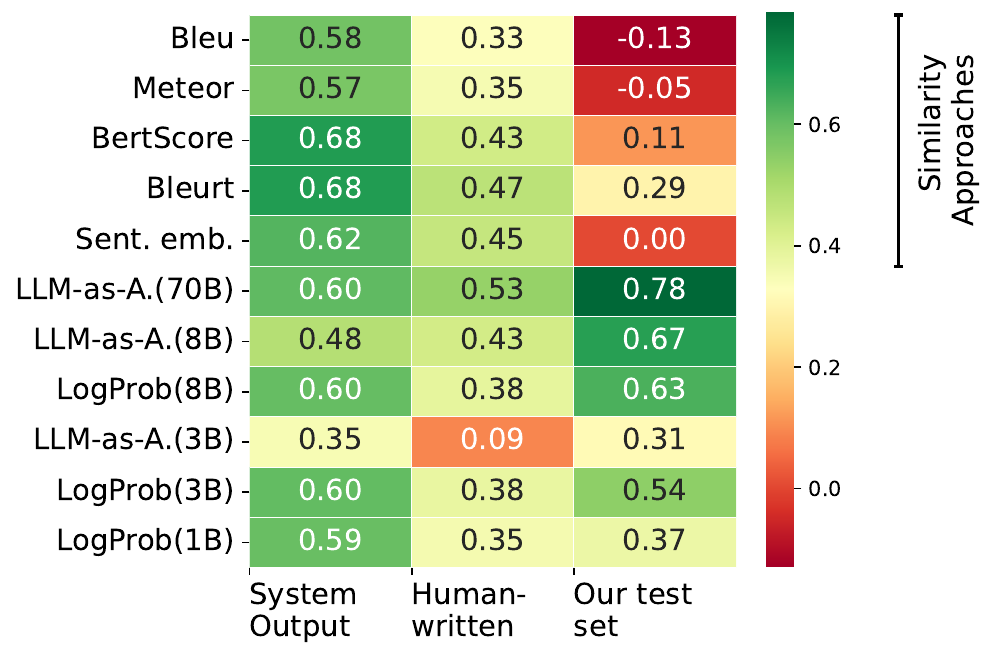}
    \caption{Average Spearman correlation between metrics and human judgments for content preservation, grouped by whether test data originates from system-generated output (total of 5 sources), human-written (total of 3 sources) or our constructed test set. Which metric is deemed best changes depending on the test data, and only our test set correctly finds similarity-based metrics to be unsuitable. 
    }
    \label{fig:rank}
\end{figure}

Fig.~\ref{fig:sample} illustrates a case where similarity-based metrics between source and output fail to correctly identify which output best preserves content, as opposed to style-aware approaches (the last two).
In this paper, we study the research question:
\textbf{Why do metrics that are theoretically unsuitable for content preservation in style transfer show a high correlation with human judgments?} as illustrated in the first two columns in Fig.~\ref{fig:rank}.

To answer this question, we conduct a large-scale meta-evaluation study with a particular focus on metrics for content preservation. 
Our key finding is that current meta-evaluation efforts for content preservation have misleading conclusions and overestimate the performance of similarity-based metrics. Consequently, the use of such metrics leads to misleading evaluation results of style transfer methods. 

We identify the data used for meta-evaluation as the primary reason for the unreliable correlation results. To overcome this data bias, we propose and construct a test set that offers carefully designed test cases  -- featuring errors such as substituted, deleted and fabricated information, which are unrelated to style, but change the content (see  Figure~\ref{fig:sample}).  

Our aggregated results (Figure~\ref{fig:rank}) demonstrate that similarity-based metrics for content preservation show low to negative correlations with human judgment on our test set (last column), but exhibit overly high correlations to human judgment on existing data sets (first two columns). Style-aware approaches show positive correlations to human judgments on all data types (all columns). Note, our dataset is designed to stress-test the metrics, and we therefore recommend assessing all data types for evaluating the metrics.

We propose the style-aware method \textbf{LogProb} to evaluate style strength and content preservation using estimated token probabilities. LogProb offers better performance on smaller language models (1B,3B) than a strong baseline of using similar-sized LLMs as autoraters. 

 In sum, our \textbf{contributions} are:
\begin{itemize}[noitemsep] 
 \item We construct a new test set\footnote{\url{github.com/AmaliePauli/style_transfer_evaluation}} (500
human-annotated samples) with carefully designed test cases as a new resource for evaluating content preservation metrics in style transfer;
\item We conduct a large meta-evaluation of content preservation spanning 9 metrics/approaches on a variety of tasks (9) and data (7), offering recommendations on metrics for measuring content preservation and style strength; 
\item We show that existing meta-evaluations may lead to misleading conclusions for metrics for content preservation, namely that widely used semantic similarity metrics are not a good fit -- we recommend to discontinue their use for evaluating style transfer;
\item We propose an efficient, zero-shot method LogProp for evaluating style transfer, adopting small language models (3B,1B), accounting for content preservation and style strength. On small model sizes, LogProb outperforms our LLM-as-Autorater baseline. 

\end{itemize}

\section{Background and Related Work}
\label{sec:background}
\textbf{Style and Attribute Transfer.} 
We use the working definition from \citet{ jin-etal-2022-deep} of \textit{style transfer as a generation task aiming to control specific attributes of the text}. Note that this definition of style transfer is data-driven and different from a linguistic definition. We likewise note that `attribute' is a broader category that covers changes that affect semantics, where a linguistic understanding of `style' would not, e.g., constitute a change of sentiment. Hence, the aim is to change the attribute/style while preserving the \textit{contextual or semantic} content.

\textbf{Evaluation.} 
Traditionally, style transfer systems have been evaluated along three dimensions: 1) style strength -- did the style successfully shift?; 2) content preservation -- is the content otherwise the same?; 3) fluency -- is the rewrite fluent and grammatically correct?  \cite{jin-etal-2022-deep}.\footnote{Note that fluency is not covered in this work} Some work on style transfer systems includes small-scale human evaluations of the systems, but papers heavily rely on automatic metrics for evaluation \cite{ostheimer-etal-2023-call}:

\textbf{Style.} 
\label{sec:back_style}
The de facto evaluation approaches, also in recent work, use either a classifier, e.g. RoBERTa, trained on a specific style \cite{ lai-etal-2024-style,hallinan-etal-2023-steer,han-etal-2024-disentangled,mukherjee-etal-2024-large-language,liu2024adaptive,luo-etal-2023-prompt,zeng2024bat}, or a regressor \cite{briakou-etal-2021-evaluating,lai-etal-2022-human,pauli-etal-2025-measuring}. However, both require training data for the specific style, limiting their scope and scale. Further, \citet{briakou-etal-2021-evaluating} criticises the use of binary classifiers, as this approach does not capture that a style can shift to differing degrees. In this paper, we focus on evaluating arbitrary styles without trained classifiers or regressors, and we focus on approaches that output scores to assess different degrees of style changes. 

\textbf{Content Preservation.} 
Many different metrics are used to evaluate style transfer for content preservation \cite{ostheimer-etal-2023-call}. The most commonly used is lexical similarity using BLEU \cite{ostheimer-etal-2023-call,jin-etal-2022-deep}, but widely used ones include BertScore, cosine distance between embeddings, and BLEURT - also in recent works \citep{lai-etal-2024-style,hallinan-etal-2023-steer,han-etal-2024-disentangled,mukherjee-etal-2024-large-language,liu2024adaptive,zhang-etal-2024-distilling,luo-etal-2023-prompt}. These metrics originate from other natural language generation tasks, e.g., translation, to measure the similarity between output and gold references.  However, in the absence of such references, many style transfer papers use these metrics to compare source and output sentences. We evaluate these approaches using source sentences compared to output [\textbf{S}ource-], and references compared to output [\textbf{R}eference-] where available. 

Prior work has criticised the use of similarity-based approaches for measuring content preservation in style transfer. \citet{mir-etal-2019-evaluating} highlight the flaw of metrics like source BLEU, which cannot distinguish between style and content changes, thus penalising style edits as lower content preservation. \citet{lai-etal-2024-style} note that similarity-based metrics favour copying over paraphrasing, disadvantaging certain transfer models. \citet{cao-etal-2020-expertise} further argue that style transfer evaluation with these metrics for content preservation can be gamed by simply appending style words to the source sentence.

A proposed solution to this criticism is to make metrics for content preservation style-aware. \citet{mir-etal-2019-evaluating} proposes to use a lexicon of style words to mask the sentence in a sentiment task. In a follow-up study,  \citet{yu-etal-2021-rethinking-sentiment} extract a style lexicon automatically. More recent work prompts an LLM for evaluation (LLM as an autorater, \citet{zeng2024bat,lai2023multidimensional,ostheimer-etal-2024-text}). For other natural language generation tasks, LLM-as-a-judge approaches are deemed better than similarity-based metrics when tasks are semantically nuanced \cite{li2024generation}.

Still, despite the established criticism of the use of semantic similarity metrics for content preservation in style transfer, these metrics often show reasonable correlation with human judgment on content preservation in style transfer. Prior meta-evaluation efforts \citep{lai2023multidimensional, ostheimer-etal-2024-text} show that using LLMs for evaluating content preservation is only on par rather than superior to using semantic similarity-based metrics. 

 \textbf{We investigate the discrepancy between what would conceptually be a suitable metric and what meta-evaluation results show.} In addition, we propose a new target-style-aware metric with improved efficiency compared to an LLM-as-a-judge approach.   

\textbf{Meta-evaluation.} 
In order to determine suitable metrics for evaluating style transfer, prior work has examined how metrics correlate with human judgments. Human judgments are often collected as multiple annotations, for style, fluency, and content preservation separately, e.g., asking how successful the style change is on a scale from 1 to 5.  

 Examples of meta-evaluation studies on different style transfers include sentiment \cite{yu-etal-2021-rethinking-sentiment,mir-etal-2019-evaluating,ostheimer-etal-2024-text}, formality \cite{briakou-etal-2021-evaluating,lai-etal-2022-human}, simplification  \cite{scialom2021rethinking, alva-manchego-etal-2021-un, cao-etal-2020-expertise}, and detoxifying \cite{logacheva-etal-2022-study}. 
  
 \paragraph{Criticism.} Prior studies have voiced different criticisms about the meta-evaluations themselves: For simplicity transfer, \citet{scialom2021rethinking} find that using system-generated rewrites for meta-evaluations leads to spurious correlation due to inter-correlation between dimensions, e.g. rewrites low on fluency also tend to be low on content and simplicity. Instead, they propose using human-written data for meta-evaluation purposes. In our broader study, we also group data into system-generated and human-written outputs. We further construct a more challenging dataset, since we argue that existing human-written test sets do not include instances of low preserved content.   Also, for a simplicity task, \citet{devaraj-etal-2022-evaluating} examine faithfulness mistakes in training data and in rewrites by systems. They show that metrics have more difficulty locating some error types related to low faithfulness -- we take inspiration from these error types when constructing our test set. \citet{logacheva-etal-2022-study} conclude, on detoxifying transfer, that automatic metrics are less reliable for high-performing systems, as the correlation between metrics and human judgment is lower for some systems' output, thereby pointing out an issue of meta-evaluation itself.

In our paper, we address the problem of meta-evaluation across a wide range of style transfer tasks. We propose a new test set, which is constructed to show the ability of metrics to detect low content preservation when the style or attribute change.

\section{Methodology: Test Set}
\label{sec:testset}
We propose a new dataset to test metrics on content preservation used in style transfer, inspired by several shortcomings we observe in existing meta-evaluation studies.  
Namely, meta-evaluation studies (both prior studies described in Section~\ref{sec:background} and our study in Section~\ref{sec:results}), show e.g. that similarity-based metrics highly correlate with human judgments and even outperform LLM-based approaches. However, we hypothesise that this is a misleading result caused by shortcomings in how the meta-evaluations are conducted.  The reason is that conceptually, similarity-based metrics for content preservation between source and output sentences are unsuitable for the task, because they favour verbatim repetitions between source and output sentences, and the more the style changes, the more the similarity drops despite the content being preserved (Section~\ref{sec:background}). 

We illustrate the shortcomings of similarity-based metrics with an example (Fig.~\ref{fig:sample}).  Suppose the task is to rewrite \textit{I don't appreciate your lack of effort} to be more polite. Consider these rewrites:
\begin{itemize}[nolistsep,noitemsep,leftmargin=*]
    \item Output 1: \textit{I would like to express my concern regarding the effort being put forth}
    \item Output 2: \textit{I would appreciate less effort in the area}
\end{itemize}
Human annotators rate style shift on a 1–5 scale (5 = very good). Both outputs are seen as successfully shifting to a more polite tone, but Output 2 scores poorly on content preservation. Yet, similarity-based metrics wrongly rate Output 2 as better at preserving content than Output 1, while style-aware metrics correctly assign it a lower score. The employed metrics are formally introduced in Section~\ref{sec:bencmarking}.

Figure~\ref{fig:sample} is an example of how we can construct more challenging data to test the abilities of content preservation metrics to detect low content preservation under style shifts. 

We aim to create a challenging test set for meta-evaluating metrics for content preservation.
Our core principle is to create data that succeeds in style transfer but has a high variation in content preservation. We create variation in content preservation by deliberately creating mistakes in the content, such that we can assess the metrics' suitability to detect low content preservation. We supply the data with human judgments for both style strength and content preservation. For each source sentence, we construct two rewrites: 
\begin{itemize}[noitemsep,nolistsep]
    \item [\textbf{1)}]one where we preserve the content, and shift the style to a large degree, with more variation in the rewrite;
    \item [\textbf{2)}] one where we shift style to a lesser degree, staying closer to the wording in the source sentence but, introducing a content error. 
\end{itemize}

The methodology for adding content errors is inspired by \citet{devaraj-etal-2022-evaluating} -- they analyse text simplification by categorising errors as substitution, deletion and insertion. For various style transfer tasks, we construct errors with these categories in mind -- we substitute or swap key information, drop key information, or fabricate additional information not supported by the source sentence.

 We aim to construct data where all output sentences succeed on style transfer,  but for each source sentence, only one output preserves the content well, creating variation in content preservation quality.
 
 Other examples from our test set include the task of rewriting the following sentence to be detoxified; \textit{yeah but she is making me feel like shit}; with the two rewritings:
 \begin{itemize}[nolistsep,noitemsep,leftmargin=*]
    \item Output 1: \textit{Yes I know, but she is really making me feel terrible}
    \item Output 2: \textit{yeah but she is making me feel}
\end{itemize} Another example is a neutral headline from WikiNews to be framed more positively; \textit{President of China lunches with Brazilian President}; with the rewritings;
\begin{itemize}[nolistsep,noitemsep,leftmargin=*]
    \item Output 1: \textit{The Great Presidents of China and Brazilian strengthen important ties over lunchs}
    \item Output 2: \textit{The President of China enjoys lunches with the Brazilian first lady}
\end{itemize}

 In total, our constructed test data covers six style/attribute tasks and consists of 500 samples, 100 of which are manually created and the remaining generated using an LLM for scale.
 
\textbf{Manually created}. We construct 100 transfer samples from a source sentence by manually creating two output sentences as described. We create: Task 1) on sentiment (positive/negative), using headlines from Wiki News\footnote{Wikinews.org} as source sentences with minor modifications; Task 2) on detoxifying, using toxic sentences from \citet{logacheva-etal-2022-paradetox}.  We supply each sample with a gold-truth reference sentence to enable a small evaluation of reference-based methods. 

\textbf{LLM generated}.  We use GPT-4o-mini\footnote{openai.com} to create style transfer pairs in a multi-step process: 
\begin{itemize}[nolistsep,noitemsep]
\setlength\itemsep{0em}
    \item[1.] Generate source sentences.
    \item[2.] Generate two rewrites for each source:
    \begin{itemize}
    \item[a.]  \textit{a lot more} in the target style
    \item[b.] \textit{a bit more} in the target style with
    a content error of substitution, deletion or fabrication of information.
    \end{itemize}
\end{itemize}

We construct the LLM-generated part of the test set with four different style transfer tasks: i) a headline to be more catchy; ii) an impolite sentence from an email to be polite; iii) a persuasive request to be more persuasive; and iv) a sentence with informal language, with grammatical mistakes and internet slang to be formal. Prompt details are included in App.~\ref{app:syn}, where we also include an automatic assessment of the `linguistic acceptability' of the samples.    

\subsection{Human Annotations}
We multi-annotate our samples and obtain a good level of agreement:
Three workers annotate each sample in batches, grouped by style task, with a total of five different workers.
Annotators rate, on a 5-point Likert scale, how well the style or attribute change is achieved, and how well the meaning/content unrelated to the style/attribute change is preserved, following previous work \cite{mir-etal-2019-evaluating,ziegenbein-etal-2024-llm}. We use the scale
\begin{lstlisting}[basicstyle=\ttfamily,
  breaklines=true]
1: Very poor, 2: Poor, 3: Fair, 4: Good, 5: Very Good.
\end{lstlisting}
We achieve a high level of human agreement on content preservation with a Krippendorff's Alpha \cite{krippendorff2011computing} of \textbf{$0.768$}, and a good level of success in style transfer, with an average score on the 5-point Likert scale of $4.27$. Detailed results per task in Table~\ref{tab:con_data}, where we report the percentage of samples that at least obtain an average rating of 3 (fair). Most sentences successfully transfer style.  

The workers are recruited via a crowdsourcing platform (prolific.com). The five different workers who participate are from the UK, hold a BA in Arts and are experienced on the platform. Workers are paid a fixed amount per batch at a rate, which the platform considers a 'great' hourly pay. Details on annotation guideline, setup and payment are provided in App.~\ref{app:anno}.
 
\begin{table}[t]
\begin{tabular}{llrlr}
\toprule
\multicolumn{1}{l}{\textbf{Task}} & \multicolumn{1}{l}{\textbf{}} & \multicolumn{1}{r}{\textbf{\#}} & \multicolumn{1}{l}{\textbf{\begin{tabular}[c]{@{}c@{}}C (IAA) \end{tabular}}} & \multicolumn{1}{c}{\textbf{\begin{tabular}[c]{@{}c@{}}S>=3 (\%)\end{tabular}}} \\ \midrule
sentiment & ma & 50 & 0.676 & 88 \\
detoxify & ma & 50 & 0.758 & 96 \\
catchy & LLM & 100 & 0.806 & 90 \\
polite & LLM & 100 & 0.645 & 100 \\
persuasive & LLM & 100 & 0.80 & 88 \\
formal & LLM & 100 & 0.817 & 99 \\ \bottomrule
\end{tabular}
\caption{Stats on our test set. ma: manually created, LLM: LLM-generated. C: content preservation, S: style strength. All samples are manually annotated by three raters, and IAA shows the inter-annotator-agreement using Krippendorff's alpha.} 
\label{tab:con_data}
\end{table}

\section{Methodology: LogProb}
\label{sec:method}

We propose a novel approach for evaluating style transfer with the following properties: 1)  no training data required on style; 2) no need for gold truth / reference sentences; 3) improved feasibility in the form of model size (e.g. 1B,3B) with improved performance compared to our baseline of prompting an LLM-as-Autorater. For content preservation, we additionally aim for: 4) accounting for changes in target style -- ensuring theoretical suitability as discussed in Section~\ref{sec:background}. 
For style strength, beyond properties 1,2 and 3, we aim for: 5) scoring on a range rather than a binary score -- motivated by the fact that styles can shift to various degrees as discussed in Section~\ref{sec:back_style} and recommended by \citet{briakou-etal-2021-evaluating}.

Similar to the approach in \citet{jia-etal-2023-zero}, who evaluate the faithfulness of summaries, we use token likelihood estimates to evaluate a style transfered sentence.  The idea in our work is to provide \textit{different} rewrite instructions as part of the context \textit{before} the rewrite sentence, e.g. with and without the target style. See Table~\ref{tab:context} for specific implementation.  This yields token likelihood estimates that vary based on whether a style change is expected. We use these estimates to measure both style strength and content preservation. Using the likelihood estimates of smaller models as opposed to generated answers offers more robustness, since prompting may result in inconsistent scores depending on the prompt, as well as inconsistent adherence to output formats, complicating postprocessing. Accessing the likelihoods avoids these issues.

Let \( X \) be the source sentence consisting of tokens \( x_1, x_2, \dots, x_m \), and let \( \tilde{X} = \tilde{x}_1, \tilde{x}_2, \dots, \tilde{x}_n \) be the rewritten sentence that we want to evaluate with respect to some style change \( S \). From an LLM, we estimate the probability of the output sentence when the model has seen the source sentence and some rewrite instruction as part of the context \( T \), denoted as \  \( P^{LM}(\tilde{X}|X,T) \). 

\paragraph{Content Preservation}
Our requirement is to measure content preservation with respect to the style change, such that the content unrelated to the style should be preserved. Note that content preservation could also be high without a style change. Hence, the measure of content preservation should be high if one of the following conditions is likely: 1)  the tokens of \(\tilde{X}\) are likely a rewrite with respect to the style $S$; or 2)  the tokens of \(\tilde{X}\) are likely a paraphrase of $X$; or 3) the special case that the tokens of \(\tilde{X}\) are likely a repetition of $X$. This implies that content \textit{is not} well preserved if the likelihood of a token is low in all three cases. Hence, if a token can neither be attributed to repetition, paraphrasing, nor requested style change, then there must be a change in the text that does not preserve the context. To weigh these cases with tokens \textit{not} preserving content, we take the log of the most likely token of our three cases to define our measure: 
\begin{align}
    C = \frac{1}{n} \sum_{i=1}^{n} \log \left( \max \left( p_{i}^{t^s}, p_{i}^{t^{pa}},p_{i}^{t^r}  \right) \right)
\end{align}
\vspace{-2em}
\begin{align*}
    p_{i}^{t^s}& = p^{LM}(x_i | X, \tilde{x}_{<i}, t^s), \\
    p_{i}^{t^{pa}}& = p^{LM}(x_i | X, \tilde{x}_{<i}, t^{pa}), \\ p_{i}^{t^{r}}& =p^{LM}(x_i | X, \tilde{x}_{<i}, t^r)
\end{align*}
and $t^s$,$t^{pa}$,$t^r$ being instructions on rewrite with respect to style, paraphrase and repeat.

\paragraph{Style} To assess how well style is transferred, we compare:
1) the likelihood of tokens in \(\tilde{X}\) after seeing a context containing an instruction to rewrite to the target style and 2) the likelihood after seeing a context containing an instruction to paraphrase or repeat, without reference to style.
The hypothesis is that tokens contributing to the target style transfer will have a higher likelihood when the context prior emphasises rewriting in this style. Thus, we measure style transfer success as the difference in token likelihood given different context instructions in `style rewrite' and `paraphrase/repetition'.
\begin{align}
    S=\frac{1}{n}\sum_{i=1}^{n} \left(p_{i}^{t^s} - \max \left(p_{i}^{t^{pa}},p_{i}^{t^r}  \right)\right)
\end{align}
We deploy three backbone model sizes—1B,3B and 8B parameters—using Llama 3 Instruct models \cite{dubey2024llama}.\footnote{ \textsc{meta-llama/Llama-3.2-3B-Instruct},\textsc{meta-llama/Llama-3.2-1B-Instruct} and \textsc{meta-llama/Llama-3.1-8B-Instruct}.} We deploy different instructions as part of the context as per Table~\ref{tab:context}. Further details in App.~\ref{app:method}.

\begin{table}[ht]
\begin{tabular}{@{}l@{}}
\toprule
\multicolumn{1}{c}{\textbf{Contexts}} \\ \midrule
$X,t^{s}$: \textit{`Rewrite the following sentence to be $S$: $X$}' \\
$X,t^{pa}$: \textit{`Paraphrase the following sentence: X'} \\
$X,t^{r}$: \textit{`Repeat the following sentence: $X$'} \\ \bottomrule
\end{tabular}
\caption{Context with different instructions on the rewrite}
\label{tab:context}
\end{table}

\section{Benchmarking}
\label{sec:bencmarking}
 Many metrics are used to evaluate content preservation for style transfer. Here, we evaluate several recently and widely used ones (per Section~\ref{sec:background}). We group content preservation metrics into four categories: lexical similarity, semantic similarity, factual-based, and style-aware. For the similarity-based metrics, if reference data is available, we test both source-based [S-](output vs. source) and reference-based [R-] (output vs. reference) approaches (For details and prompts, see App.~\ref{app:method}):
\begin{itemize}[noitemsep,leftmargin=*]
    \item {\ul{Lexical similarity:}} \textbf{BLEU} \cite{papineni-etal-2002-bleu} and \textbf{Meteor} \cite{banerjee-lavie-2005-meteor} -- both are n-gram match metrics originally developed for evaluating machine translation. 
    \item  {\ul{Semantic similarity:}} token-based \textbf{BERTScore} \cite{zhang2019bertscore}; \textbf{Cosine} similarity between sentence embeddings \cite{feng-etal-2022-language}; learned BERT-based \textbf{BLEURT} \cite{sellam-etal-2020-bleurt}. \textbf{COMET} \cite{li2024generation}, which uses both source and references to evaluate; however, COMET also have a reference-free version. 
    \item {\ul{Fact-based:}} \textbf{QuestEval} \cite{scialom-etal-2021-questeval}: uses question generation and question answering to evaluate answers on source and output; originally for factual consistency and relevance in summaries, later extended to faithfulness in simplification tasks \cite{scialom2021rethinking}.  
    \item {\ul{Style-aware:}} \textbf{LLM-as-Autorater}: Llama3 Instruct \cite{dubey2024llama} as an evaluator: prompting for a score on a 5-point scale given source sentence, style and output sentence for both content-preservation and style strength (70b parameter and 3b,8b);  \textbf{LogProb}: our method using likelihood estimates, per Section~\ref{sec:method}. 
\end{itemize}
 We assess evaluation approaches on style strength using \textbf{LLM-as-Autorater} and our method \textbf{LogProb}, both of which as zero-shot approaches.
 
\paragraph{Datasets} We benchmark nine metrics for evaluating content preservation in style transfer on seven different datasets covering nine styles. We use existing datasets already containing human evaluations of style strength and/or content preservation. The human evaluations are conducted on rewrites from system-generated output and/or human-written / gold reference output, using various scales (e.g., 5-point Likert or 1–100). These datasets cover rewriting tasks for simplifying, sentiment, formality, and making arguments more appropriate. The following datasets are used (abbreviation in brackets), in addition to our newly test set (Section~\ref{sec:testset}) -- further details in App.~\ref{app:dataset}:
\begin{itemize}[noitemsep,leftmargin=*]
    \item \citet{lai-etal-2022-human} \textbf{[Lai]} formal/informal,
    \item \citet{mir-etal-2019-evaluating} \textbf{[Mir]}  positive/negative,
    \item \citet{alva-manchego-etal-2020-asset} \textbf{[Alv.]} simplifying, 
    \item \citet{scialom-etal-2021-questeval} \textbf{[Sci.]}  simplifying,
    \item \citet{ziegenbein-etal-2024-llm} \textbf{[Zie.]} appropriated arguments,
    \item \citet{cao-etal-2020-expertise} \textbf{[Cao]} layman/expert.
\end{itemize}

We report Spearman rank correlation between metric scores and mean human judgement.
We group test data as rewrites generated by systems, human-written gold-references, and our constructed test set.
We summarise results in each group by reporting the average Spearman correlation across data sources (Avg), and the average ranking of the metric per data source by correlation score (Avg. rank).


\begin{table*}[]
\centering
\fontsize{10pt}{10pt}\selectfont
\begin{tabular}{@{}llllllll|lllll@{}}
\toprule
\multicolumn{8}{c|}{\textbf{System-Generated ouput}} & \multicolumn{5}{c}{\textbf{Gold-reference output}} \\ \midrule
\textbf{} & \textbf{Mir} & \textbf{Lai} & \textbf{Zei.} & \textbf{Cao.} & \textbf{Alv.} & \textbf{avg} & \textbf{\begin{tabular}[c]{@{}l@{}}avg \\ rank\end{tabular}} & \textbf{Lai} & \textbf{Zei} & \textbf{Sci.} & \textbf{avg.} & \textbf{\begin{tabular}[c]{@{}l@{}}avg. \\ rank\end{tabular}} \\ \midrule
\rowcolor[HTML]{E8E8E8} 
\textbf{S-Bleu} & 0.51 & 0.52 & 0.63 & 0.58 & 0.66 & 0.58 & 6.4 & 0.44 & 0.38 & 0.17* & 0.33 & 9 \\
\rowcolor[HTML]{E8E8E8} 
\textbf{R-Bleu} & -- & 0.2 & 0.26 & -- & -- & -- &  &  &  &  &  &  \\
\textbf{S-Meteor} & 0.49 & 0.48 & \textbf{0.65} & 0.59 & 0.64 & 0.57 & 6.9 & 0.45 & 0.44 & 0.15* & 0.35 & 8.17 \\
\textbf{R-Meteor} & -- & 0.24 & 0.24 & -- & -- & -- &  &  &  &  &  &  \\
\rowcolor[HTML]{E8E8E8} 
\textbf{S-BertScore} & \textbf{0.52} & \textbf{0.67} & 0.74 & {\ul 0.65} & {\ul 0.81} & {\ul 0.68} & \textbf{1.6} & {\ul 0.54} & 0.44 & 0.31 & 0.43 & 4.8 \\
\rowcolor[HTML]{E8E8E8} 
\textbf{R-BertScore} & -- & 0.36 & 0.38 & -- & -- & -- &  &  &  &  &  &  \\
\textbf{S-Bleurt} & {\ul 0.51} & \textbf{0.67} & \textbf{0.65} & \textbf{0.71} & \textbf{0.86} & \textbf{0.68} & {\ul 2.0} & 0.48 & {\ul 0.52} & {\ul 0.41} & {\ul 0.47} & 3.7 \\
\textbf{R-Bleurt} & -- & 0.4 & -0.05* & -- & -- & -- &  &  &  &  &  &  \\
\rowcolor[HTML]{E8E8E8} 
\textbf{S-Cosine} & \textbf{0.52} & 0.57 & \textbf{0.65} & 0.61 & 0.74 & 0.62 & 4.0 & \textbf{0.61} & 0.48 & 0.26 & 0.45 & {\ul 3.5} \\
\rowcolor[HTML]{E8E8E8} 
\textbf{R-Cosine} & -- & 0.27 & 0.27 & -- & -- & -- &  &  &  &  &  &  \\
\textbf{S-Comet} & 0.21 & 0.31 & -0.13 & 0.04 & 0.31 & 0.15 & 12.8 & 0.01* & 0.19 & 0.09* & 0.10 & 12.0 \\ 
\textbf{RS-Comet} & -- & 0.46 & 0.25 & -- & -- & -- &  &  &  &  &  &  \\ 
\midrule
\rowcolor[HTML]{E8E8E8} \textbf{QuestEval} & 0.26 & 0.2 & 0.49 & 0.48 & 0.61 & 0.41 & 11.3 & 0.25 & 0.35 & 0.01* & 0.20 & 11.7 \\
\textbf{LLM-as-A 70b} & 0.35 & 0.66 & 0.63 & 0.63 & 0.75 & 0.60 & 4.7 & 0.56 & \textbf{0.57} & \textbf{0.46} & \textbf{0.53} & \textbf{1.3} \\
\textbf{LLM-as-A 8b} & 0.3 & 0.5 & 0.38 & 0.54 & 0.67 & 0.48 & 9.5 & 0.45 & 0.5 & 0.33 & 0.43 & 4.8 \\
\textbf{LLM-as-A 3b} & 0.26 & 0.44 & 0.11 & 0.47 & 0.45 & 0.35 & 12.3 & 0.33 & 0.15 & -0.21* & 0.09 & 11.7 \\
\rowcolor[HTML]{E8E8E8} 
\textbf{LogProb 8b} & 0.48 & 0.64 & 0.57 & 0.56 & 0.74 & 0.60 & 6.8 & 0.49 & 0.48 & 0.18* & 0.38 & 5.5 \\
\rowcolor[HTML]{E8E8E8} 
\textbf{LogProb 3b} & 0.49 & 0.65 & 0.58 & 0.56 & 0.73 & 0.60 & 6.3 & 0.51 & 0.47 & 0.15* & 0.38 & 6.3 \\
\rowcolor[HTML]{E8E8E8} 
\textbf{LogProb 1b} & 0.5 & 0.65 & 0.56 & 0.54 & 0.71 & 0.59 & 7.0 & 0.51 & 0.42 & 0.12* & 0.35 & 7.8 \\ \bottomrule
\end{tabular}
\caption{Spearman rank correlation between metric and human judgment split on data from system-generated or gold-reference rewrites. `R-': reference-based, `S-': source-based. Average correlation across datasets [avg.], average rank of metrics per dataset [avg. rank]. *not significant (level 0.05). Best metric per dataset in bold; second best underlined.}
\label{tab:syshum}
\end{table*}
\begin{table*}[]
\centering
\fontsize{10pt}{10pt}\selectfont
\begin{tabular}{@{}lllllllll@{}}
\toprule
 & \textbf{ALL} & \textbf{Sentiment} & \textbf{Detoxify} & \textbf{Catchy} & \textbf{Polite} & \textbf{Persuasive} & \textbf{formal} & \textbf{avg rank} \\ \midrule
\rowcolor[HTML]{E8E8E8} 
\textbf{S-Bleu} & -0.13 & -0.79 & -0.51 & -0.14* & -0.07* & -0.15* & -0.1* & 12.7 \\
\rowcolor[HTML]{E8E8E8} 
\textbf{R-Bleu} &  & -0.49 & -0.51 & -- & -- & -- & -- &  \\
\textbf{S-Meteor} & -0.05* & -0.6 & -0.44 & -0.02* & 0.06* & -0.13* & -0.03* & 11.6 \\
\textbf{R-Meteor} &  & -0.18* & -0.43 & -- & -- & -- & -- &  \\
\rowcolor[HTML]{E8E8E8} 
\textbf{S-BertScore} & 0.11 & -0.44 & -0.19* & -0.22* & 0.35 & 0.05* & 0.26* & 9.3 \\
\rowcolor[HTML]{E8E8E8} 
\textbf{R-BertScore} & -0.15* & -0.33 & -- & -- & -- & -- &  \\
\textbf{S-Bleurt} & 0.29 & 0.06* & 0.37 & -0.02* & 0.33 & 0.22 & 0.46 & 7.3 \\
\textbf{R-Bleurt} &  & 0.19* & 0.29 & -- & -- & -- & -- &  \\
\rowcolor[HTML]{E8E8E8} 
\textbf{S-Cosine} & 0.00* & -0.5 & -0.3 & -0.23* & 0.1* & -0.07* & 0.17* & 11.0 \\
\rowcolor[HTML]{E8E8E8} 
\textbf{R-Cosine} &  & -0.17* & -0.28* & -- & -- & -- & -- &  \\
\textbf{S-Comet} & 0.29 & 0.4 & 0.44  & 0.29  & 0.17* & 0.33 & 0.17* & 6.8 \\
\textbf{RS-Comet} &  & -0.17* & 0.01* & -- & -- & -- & -- &  \\ \midrule
\rowcolor[HTML]{E8E8E8} \textbf{QuestEval} & 0.22 & 0.3 & 0.26* & 0.36 & 0.19* & 0.32 & 0.11* & 6.6 \\
\textbf{LLM-as-A. 70b} & \textbf{0.78} & \textbf{0.72} & \textbf{0.69} & \textbf{0.81} & \textbf{0.82} & \textbf{0.82} & \textbf{0.74} & \textbf{1.0} \\
\textbf{LLM-as-A. 8b} & {\ul 0.67} & {\ul 0.41} & {\ul 0.64} & {\ul 0.75} & 0.64 & {\ul 0.67} & 0.7 & {\ul 2.5} \\
\textbf{LLM-as-A. 3b} & 0.31 & 0.29 & 0.24* & 0.45 & 0.19* & 0.39 & 0.34 & 6.3 \\
\rowcolor[HTML]{E8E8E8} 
\textbf{LogProb 8b} & 0.63 & 0.12* & 0.49 & 0.72 & {\ul 0.7} & 0.51 & {\ul 0.72} & 3.2 \\
\rowcolor[HTML]{E8E8E8} 
\textbf{LogProb 3b} & 0.54 & -0.08* & 0.32 & 0.59 & 0.6 & 0.38 & 0.71 & 5.0 \\
\rowcolor[HTML]{E8E8E8} 
\textbf{LogProb 1b} & 0.37 & -0.18* & 0.11* & 0.32 & 0.51 & 0.23 & 0.52 & 7.2 \\ \bottomrule
\end{tabular}
\caption{Spearman correlation between metrics and human judgment on \textbf{our testset}. `R-': reference-based, `S-': source-based. Correlation on entire testset [ALL], average rank per dataset [avg. rank]. *not significant (cannot reject null hypothesis of zero correlation) level 0.05. Bold best metric per dataset; second best underlined. }
\label{tab:con}
\end{table*}

\begin{table*}[]
\centering
\fontsize{10pt}{10pt}\selectfont
\begin{tabular}{@{}lllllll@{}}
\toprule
 & \textbf{Mir} & \textbf{Lai} & \textbf{Zei.} & \textbf{Alv.} & \textbf{Sci.} & \textbf{avg.} \\ \midrule
\textbf{LLM-as-A. 70b} & 0.45 & \textbf{0.6} & 0.28 & \textbf{0.6} & 0.39 & 0.46 \\

\textbf{LLM-as-A 8b} & 0.34 & {\ul 0.55} & 0.18 & {\ul 0.32} & 0.2* & 0.32 \\
\textbf{LLM-as-A. 3b} & 0.29 & 0.37 & 0.14 & 0.23 & 0.01* & 0.21 \\
\rowcolor[HTML]{D9D9D9} 
\textbf{LogProb 8b} & \textbf{0.5} & 0.28 & \textbf{0.32} & -0.02* & \textbf{0.5} & \cellcolor[HTML]{D9D9D9}0.32 \\
\rowcolor[HTML]{D9D9D9} 
\textbf{LogProb 3b} & {\ul 0.46} & 0.28 & {\ul 0.29} & -0.01* & 0.45 & 0.29 \\
\rowcolor[HTML]{D9D9D9} 
\textbf{LogProb 1b} & 0.41 & 0.31 & 0.27 & -0.03* & {\ul 0.48} & \cellcolor[HTML]{D9D9D9}0.29 \\ \bottomrule
\end{tabular}
\caption{Style strength: Spearman correlation between metrics and human ratings. *not significant (level 0.05)}
\label{tab:styleresults}
\end{table*}

\section{Results}
\label{sec:results}

\textbf{Meta-evaluation is highly impacted by the type of test samples.} Overall metric performance (correlation with human judgements) varies greatly depending on whether the human-rated outputs are system-generated, human-written, or from our newly constructed test set (Fig.~\ref{fig:rank}).
    \textbf{On system-generated data} (Table~\ref{tab:syshum}): Many metrics show relatively high correlations across tasks. For all similarity-based metrics except COMET, the versions using source sentences [S-] result in higher correlations to human ratings than the versions using references [R-].  The best metrics overall are S-BertScore and S-BLEURT (avg+avg rank). 
    \textbf{On gold reference data} (Table~\ref{tab:syshum}): Here, LLM-as-Autorater 70B performs best (avg+avg rank), but semantic-similarity based metrics still perform second best (S-BertScore and S-BLEURT). The ranking of best metrics changes when the underlying test data with human ratings originates from human-written/gold reference outputs or system-generated outputs. This is also discussed by \citet{scialom-etal-2021-questeval} for simplicity transfer. This is likely due to the nature of the system-generated output used, where verbatim repetitions happen to correlate with human judgments of content preservation. 
    \textbf{Our test set} (Table~\ref{tab:con}): Several metrics show low or no correlation, or even significant negative correlation for some metrics on the entire test set (BLEU) or on parts  (Meteor, BertScore, Cosine).

\textbf{Results on our test set show that similarity-based metrics for content preservation are unsuitable: }
 Similarity-based metrics are deemed unsuitable for content preservation in style transfer by prior work (Section~\ref{sec:background}), and we can now support this claim by assessing the correlation of these metrics to human judgments on our test set.  On our test set, similarity-based metrics show low or negative correlations with human judgment. We conclude that the high correlation scores previously reported on similarity-based metrics obtained on existing dataset are misleading. We hypothesise that misleading conclusions from prior meta-evaluations stem from the nature of the source data used: system-generated outputs happen to be very similar to source sentences lexically and semantically, whereas human-written/gold reference outputs do not, or at least rarely, contain samples where content not related to the style shift is poorly preserved or even fabricated -- both limiting the cases where the metrics ability are tested.  

We recommend that meta-evaluation efforts for content preservation go beyond measuring correlations with human judgments, and also account for the data source bias underlying those judgments. Specifically, content preservation should be benchmarked on diverse data sources -- including those with intentionally low content preservation, such as our test set. 

\textbf{{Content preservation: }} Content preservation metrics must be style-aware.
Across all tasks and data sources, overall, the highest correlations between metric and human judgement are achieved by using LLM-as-Autorater using 70B Llama 3 Instruct. If compute is a constraint, we recommend the LogProb method with a smaller model. Both model sizes of 3B and 1B outperform 3B and 1B LLM-as-Autorater \footnote{The LLM-as-Autorater 1b result is not reported as for over 50\%  of the cases, the output was not compliant with the expected output format} (Figure~\ref{fig:rank}). For a backbone model size of 8b, LLM-as-Autorater is recommended.  

\textbf{Style: }
LLM-as-Autorater 70B shows the best average performance in terms of correlations with human judgements. Still, our more efficient LogProb method scores higher than LLM-as-Autorater 70B on 3 out of 5 datasets, despite using a much smaller model (both 8B and 3B versions) (Table~\ref{tab:styleresults}). The variation in results across datasets calls for a task-specific investigation of the best metrics. 

\section{Conclusion}
\label{sec:conclusion}
We conduct a large-scale meta-evaluation of metrics for style transfer, focusing on content preservation. We construct a new challenging test set for assessing metrics for content preservation with human judgment.  We show that the meta-evaluation of metrics for content preservation is not straightforward.
Previous meta-evaluation studies on existing datasets consisting of either automatically generated data from specific systems or human-generated data alike leads to misleading conclusions about best metrics, due to skewed test cases. We hypothesise that on existing system-generated data, human judgment on content preservation happens to correlate with verbatim repetitions, and on gold references/human-written data, the limits of the metrics are not tested, as cases of drastically low preserved or fabricated content are not or rarely present in the data.
With our new test set, deliberately designed with variation in content preservation, we demonstrate empirically that similarity-based metrics are not suitable for content preservation for style transfer. 
Instead, metrics for content preservation should be style-aware. We recommend for meta-evaluation that a diverse set of data sources be used, including both system-generated test cases with human ratings, as well as our newly constructed test set, which tests the nature of content preservation metrics. 

We propose a new efficient style-aware metric for content preservation and one for style strength, reusing the same computational effort (LogProb). 

Overall, we find that a large LLM-as-Autorater (70B) achieves the highest correlation between content preservation scores and human judgments. However, if considering computational efficiency or feasibility in the form of a smaller model, then our method LogProb outperforms a LLM-as-Autorater baseline using a 3B parameter LLM.

 \section{Limitations}
 In evaluating style transfer, many different metrics are used for content preservation, leading to a need for standardization \cite{ostheimer-etal-2023-call}. While there are more metrics we could have tested, we do test the more widely used ones, as well as metrics of different types: lexical similarity, semantic similarity, fact-based, and LLMs conditioned on style shift (style-aware).  Especially for testing LLMs within the LLM-as-a-Judge paradigm, our work is limited to testing one prompting approach, as our main focus has been to establish better meta-evaluation practices regarding content preservation in style transfer. We refer to a survey of different trends in the LLM-as-Judge paradigm in \cite{li2024generation}. 
 We provide general recommendations on metrics over a large variety of rewriting tasks; however, specific tasks may have specific requirements that are not covered in this paper. In this paper, we have not covered the evaluation of `fluency' in rewrites, but mainly focused in-depth on content preservation and partly on style strength. 
\section*{Acknowledgements}
$\begin{array}{l}\includegraphics[width=1cm]{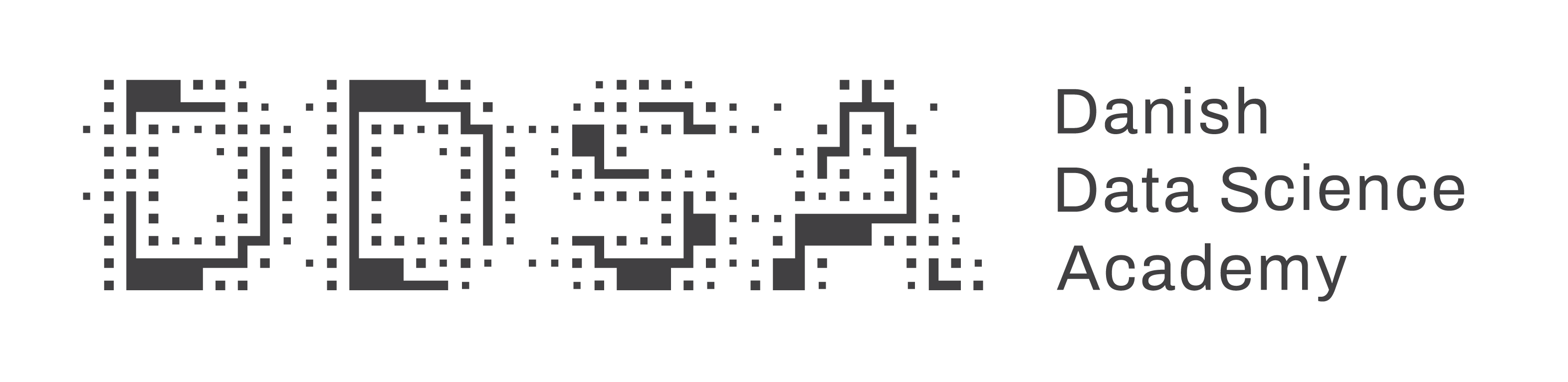}\includegraphics[width=1cm]{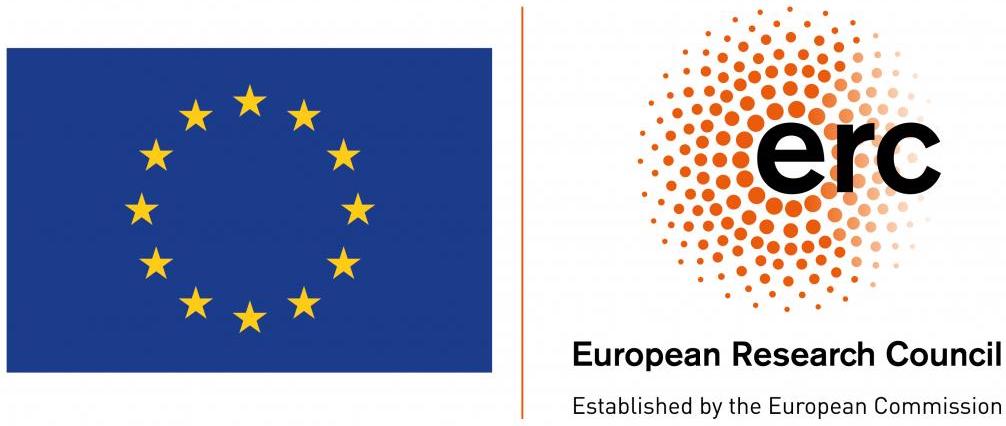} \end{array}$ This work was supported by the Danish Data Science Academy, which is funded by the Novo Nordisk Foundation (NNF21SA0069429) and VILLUM FONDEN (40516). 
It was further supported by the European Union (ERC, ExplainYourself, 101077481), and by the Pioneer Centre for AI, DNRF grant number P1.

\bibliography{custom}

\appendix

\appendix
\section{Data sets}
\label{app:dataset}
In table~\ref{tab:dataset}, we list the stats of datasets with human annotations used for benchmarking in this paper. 

The data from \citet{mir-etal-2019-evaluating} [Mir] is on the Yelp sentiment task and is annotated by 3 workers, where the mean ratings are released. Data is downloaded from  \url{github.com/passeul/style-transfer-model-evaluation}. No licence.
\\

The data from \citet{lai-etal-2022-human} [Lai] supplied human annotations on system output on the formality task, using a continuous scale from 1-100. Download at \url{https://github.com/laihuiyuan/eval-formality-transfer} with MIT License. 
\\

The data from \citet{scialom2021rethinking} [Sci] is on human ratings for human-written output for a simplification sentence task. It complements the dataset from  \citet{alva-manchego-etal-2020-asset}. Download using the URL in the paper. No license specified. We have filtered this data to obtain annotation in all three dimensions for the same data input (we check for an exact match on source sentence, rewrite, sentenceID), and we ended up with 65 samples annotated by 25 workers. 
\\

The data from \citet{alva-manchego-etal-2020-asset} [Alv.] is system output on a simplification task. We use the resource released along with \citet{scialom2021rethinking}, as it contains more metadata, such as system information and more annotations. We filter the data such that we have 135 samples with 11 annotations in all three dimensions because we favour more samples over the number of annotations per sample. 
\\

The data from \citet{ziegenbein-etal-2024-llm} [Zie.] is on rewriting inappropriate arguments to appropriate, download available at \url{https://github.com/timonziegenbein/inappropriateness-mitigation}.
\\

The data from \citet{cao-etal-2020-expertise} [Cao] is a human evaluation on a task of transferring different styles of expertise in the medical domain. The authors have kindly shared the data with human ratings. 

\begin{table*}
\centering
\small
\begin{tabular}{@{}lllllllll@{}}
\toprule
\textbf{Abb.} & \textbf{Style} & \textbf{Dim} & \textbf{Support} & \textbf{\#Ann.} & \textbf{\#Sys.} & \textbf{Ref.} & \textbf{\begin{tabular}[c]{@{}l@{}}Rating \\ on ref.\end{tabular}} & \multicolumn{1}{c}{\textbf{Scale}} \\ \midrule
Lai & \begin{tabular}[c]{@{}l@{}}formal,\\ informal\end{tabular} & S,C,F & 640 & 2 & 8 & \Checkmark & \Checkmark & 1-100 \\
Mir & \begin{tabular}[c]{@{}l@{}}positive, \\ negative\end{tabular} & S,C & 2928 & 1* & 12 &  &  & 1-5 \\
Alva-M. & simplifying & S,C,F & 135 & 11 & 6 &  &  & 0-100 \\
Scialom & simplifying & S,C,F & 65 & 25 & 1 &  & \Checkmark & 0-100 \\
Ziegen. & \begin{tabular}[c]{@{}l@{}}appropriated \\ arguments\end{tabular} & S,C,F & 1350 & 5 & 6 & \Checkmark & \Checkmark & 1-5 \\
Cao & \begin{tabular}[c]{@{}l@{}}layman, \\ expert\end{tabular} & C & 3800 & 1 & 5 &  &  & 1-5 \\ \bottomrule
\end{tabular}
\caption{Stats on the dataset used for benchmarking style transfer metrics: \textbf{Dim}ensions which of (\textbf{S}tyle, \textbf{C}ontent preservation, \textbf{F}luency) are rated in the data, \textbf{Support}: numbers of rated samples, \#\textbf{Ann}otators: number of annotations per sample, \#\textbf{Sys}stems: number of different systems/settings (including references) used to produce the samples, \textbf{Ref}erence: is the data supplied with references to enable reference-based evaluation, \textbf{Rating on ref}erence: does the data have ratings on references. *Mir dataset is conducted with multiple annotators, but only the mean of the annotations is released.}
\label{tab:dataset}
\end{table*}

\section{LLM-generated Test Part}
\label{app:syn}
A part of our construct test set is synthetically generated using GPT-4o-mini from OpenAI. We display the prompts for obtaining the samples from our stepwise approach; we show it for the subpart of politeness:
\paragraph{Generating source data} 

prompt =\textit{'Please give me} \{number\} \textit{examples of impolite sentences from emails, and return only in json format with key "sentences"'}

\paragraph{Generate rewrites}
\begin{itemize}
    \item[1)]  \textit{"Please rewrite the following} \{number\} \textit{sentences to be very polite, return in JSON with key  'sentences:'}  \{list of sentences\}"
    \item[2)]  \textit{"Please rewrite the following} \{number\} \textit{sentences to be just a bit more polite, return in JSON with key  'sentences:'}  \{list of sentences\}"
\end{itemize}
\textbf{Generate one content error in the 2. rewrite}
\begin{itemize}
    \item[a)] "\textit{Please rewrite the following } \{number\} \textit{sentences staying as close to the original wording as possible but make a mistake in the content by substituting some key information, return in json with key  'sentences:'} \{list of sentences\}
\item[b)]  \textit{"Please rewrite the following }\{number\} \textit{sentences staying as close to the original wording as possible but make a mistake in the content by omitting some key information, return in json with key  'sentences:'} \{list of sentences\}
\item[c)]  \textit{"Please rewrite the following} \{number\} \textit{sentences staying as close to the original wording as possible but make a mistake by adding a very short extra detail, return in json with key  'sentences:'} \{list of sentences\}
\end{itemize}

\paragraph{Automatic assessment of `linguistic acceptability'}
To assess the quality of the LLM-generated test samples compared to the manually created ones, we assess the `linguistic acceptability' of the samples. We use a pretrained binary classifier from HuggingFace.com  [textattack/roberta-base-CoLA],  trained on the CoLA dataset on grammatical acceptability \cite{warstadt2019neural}. We report the percentage of the samples predicted as 'linguistically acceptable' grouped by LLM-generated and manually created sampels. We observe that most samples are predicted as grammatically acceptable, Table~\ref{app:tab:ling}.

\begin{table}[t]
\centering
\small
\begin{tabular}{@{}lll@{}}
\toprule
         & \textbf{Manually created} & \textbf{LLM-generated} \\ \midrule
\textbf{source}   & 94 \%            & 93.5 \%       \\ 
\textbf{rewrites} & 80 \%            & 98.75 \%      \\ \bottomrule
\end{tabular}
\caption{Percentage of samples predicted as lingusitic acceptable}
\label{app:tab:ling}
\end{table}

\section{Annotation Guide and Process}
\label{app:anno}
\subsection{Annotation Procedure}
We recruit annotators through the crowd-sourced platform prolific.com. We use workers who we have experienced delivering high-quality in a previous annotation study. In total, we use five different workers, all from the UK and holding a BA in Arts. The workers are paid a fixed amount per task, which Prolific considers a 'great' hourly pay. Considering the completion times, the average hourly pay to the workers was $16.2$ GPB.  

One of our subtasks involves detoxifying toxic content; we warn the workers of potentially disturbing content before they select the task.

\begin{figure}
    \centering  \includegraphics[width=\linewidth]{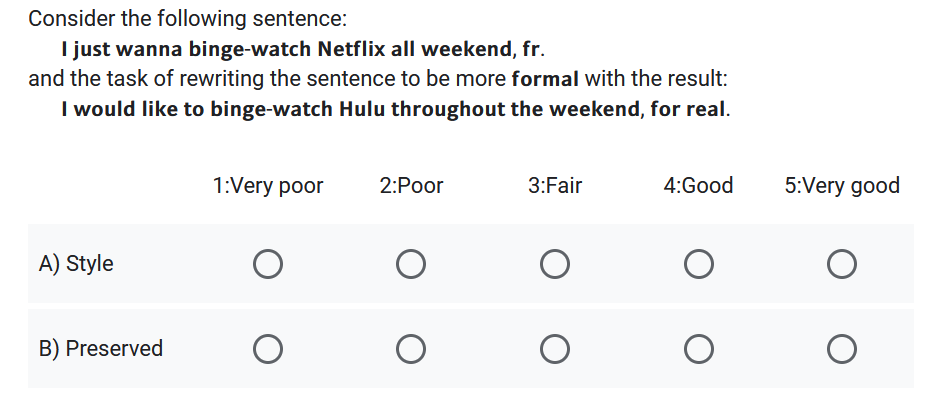}
    \caption{Screenshot of annotation tool, Google Forms}
    \label{fig:screenl}
\end{figure}

Full annotation guidelines are below, and a screenshot of the annotation interface with a sample is in Figure~\ref{fig:screenl}. We use Google Forms as our annotation tool. 

\subsection{Annotation Guideline}
\textbf{Evaluating Rewrites to Change Style/Attribute}

In this task, your goal is to help us evaluate sentences that have been paraphrased or rewritten to modify a specific style or attribute. These attributes may include tone, formality, positivity, politeness, or personalisation, among others. For example, a sentence might be rewritten to add a positive sentiment or to simplify the text for a younger audience.

When a text is rewritten to modify its style or attribute, it is important that the original content unrelated to the intended change remains intact. For example, no important information should be dropped, new information should not be fabricated, and the rewritten sentence should not mix up or misrepresent facts—except where necessary to achieve the requested change.

We will provide you with an original sentence and a rewritten version, along with the intended style or attribute change. Note that the style/attribute may vary across different examples during the study, so please pay attention to the provided description for each pair.

For each sentence pair, you will answer two evaluation questions using a 5-point Likert scale from 'Very poor' to 'Very good':
\begin{itemize}
    \item [B)] Style/Attribute Change: Evaluate how well the intended style or attribute change is achieved in the rewritten sentence.
    \item [A)] Content Preservation: Evaluate how well the meaning/content unrelated to the style or attribute change is preserved in the rewritten sentence.
\end{itemize}
The study will provide you with a link to a Google Form, where there will be 100 samples to evaluate. The study is estimated to take 35 minutes.

The annotation will be part of a research project for my PhD.

Thank you for considering participating. 

\section{Methods Implementation}
\label{app:method}
\subsection{Our method: LogProb}
\label{impl:logpob}
We deploy the backbone models \textsc{meta-llama/Llama-3.2-1B-Instruct} and
 \textsc{meta-llama/Llama-3.2-3B-Instruct} and \textsc{meta-llama/Llama-3.1-8B-Instruct} downloaded from \url{https://huggingface.co/meta-llama}. 
 
 We set the system prompt to \textit{"You can repeat sentences, paraphrase sentences or rewrite sentences to change the style or certain attribute of the text while preserving non-related content and context. Your answers contain just the rewrite."}. 

We run Python 3.13.2, and use the Transformer (4.51.3) library from Huggingface \url{https://huggingface.co/docs/transformers/index}.
We have conducted the experiments on a machine with the following characteristics:

\begin{lstlisting}
Intel Xeon Silver 4410T 
    (40) @ 2.700GHz
NVIDIA GeForce RTX 4090 2 STK
252GB RAM,
\end{lstlisting}
running Ubuntu 22.04.5 LTS.
Using this machine with one GPU, the average inference time to evaluate one sample is:
\begin{itemize}[noitemsep]
    \item \textbf{0.025} seconds per sample using the \textbf{1B} backbone model
    \item \textbf{0.041} seconds per sample using the \textbf{3B} backbone model
    \item \textbf{0.070 }seconds per sample using the \textbf{8B} backbone model
\end{itemize}
calculated from time to evaluate our test set of 500 samples.

\subsection{LLM-as-Autorater}

We deploy LLM-as-Autorater using both the 70B parameter model and 8B,3B, and 1B versions of the Llama 3 Instruct models.

For the 70B version, we use API calls to \url{replicate.com}, and the model \textsc{meta-llama-3-70b-instruct} (\url{replicate.com/meta/meta-llama-3-70b-instruct}). We use the hyperparameter settings "temperature": 0.5, "top\_p": 0.9,  else default. 

For the 1B,3B and 8B versions, we use the same machine as described in Section~\ref{impl:logpob}.
We set hyperparameters in order to lower computation at inference time: "temperature"=1.0,"do\_sample"=False, "max\_new\_tokens"~=20.  The 1B version did not produce results that were compliant with the requested output format. The inference times for the 3B and 8B versions are:
\begin{itemize}[noitemsep]
    \item \textbf{0.20} seconds per sample using the \textbf{3B} backbone model
    \item \textbf{0.37 }seconds per sample using the \textbf{8B} backbone model
\end{itemize}
calculated from time to evaluate our test set of 500 samples. 

We use the following prompt for the 70b version:
\begin{itemize}
    \item \textit{"Evaluate the following completion of a task where a 'source sentence' has been rewritten to be more} \{style\} i\textit{n the style, denoted 'target sentence', Ideally the context and content in the sentence which does not relate to the style should be preserved. Please evaluate on a Likert scale from 1-5 with 5 being the best: 1) how well the meaning is preserved and 2) how well the style is changed. Return in JSON format with the keys 'meaning' , 'style'. Given the 'source sentence':} \{source sentence\} \textit{'target sentence':} \{rewrite\}"
\end{itemize}
On the 8b and 3b backbone model, we add an 'only' in the sentence 'Return only JSON format...'.
 It occurs rarely that the LLM does not provide an answer in the right format. In these cases, we provide the mean rating of the datasets. The number of occurrences where this happens is 0.3 \% for the 70b, 0 \% for the 8b, and 0.12 \% for the 3b backbone model.  
 
\subsection{Metrics}
\textbf{BLEU} \cite{papineni-etal-2002-bleu} we use the python package NLTK (3.9.1) implementations of BLEU with default settings.

\noindent \textbf{Meteor} \cite{banerjee-lavie-2005-meteor} we use the python package from Huggingface Evaluate (0.4.3) with default settings. 

\noindent \textbf{BertScore} \cite{zhang2019bertscore} We use the implementation from \url{https://github.com/Tiiiger/bert_score} (MIT license) with the current recommended backbone model \textsc{microsoft/deberta-xlarge-mnli}.

\noindent \textbf{BLEURT} \cite{sellam-etal-2020-bleurt} we use the python implemention from \url{https://huggingface.co/Elron/bleurt-large-512} using Huggingface Transformer libary with the backbone model \textsc{Elron/bleurt-large-512}.

\noindent \textbf{Cosine similarity embeddings} we use the SentenceTransformer (2.7.0) library with Labse embeddings \textsc{sentence-transformers/LaBSE}, \cite{feng-etal-2022-language}.

\noindent \textbf{QuestEval} we use the implementations from \url{https://github.com/ThomasScialom/QuestEval} (MIT license).

\noindent \textbf{COMET} \cite{rei-etal-2020-comet} we use the python package from Huggingface Evaluate (0.4.3) with default settings for the reference and source version [RS-COMET]. We use the guide and Python implementation from \url{https://github.com/Unbabel/COMET} with the recommended model [Unbabel/wmt22-cometkiwi-da] for reference-free evaluation.  

\section{Fluency} The dimension of fluency is not covered in this work of meta-evaluating metrics for style transfer, as the main focus has been on content preservation and the discrepancy between what are suitable metrics and what the empirical results show.  Furthermore, one of the main strengths of LLMs is their ability to generate very fluent sentences. However, we include a small assessment of fluency using the datasets, which has annotations on this dimension. We deploy two commonly used methods: perplexity using GPT2 \cite{radford2019language} and classification on grammatical acceptability. To calculate perplexity using GPT2 (ppl), we utilise the Huggingface Library and the model available on [openai-community/gpt2]. To predict grammatical acceptability (clf\_cola), we use a pretrained binary classifier from HuggingFace.com  [textattack/roberta-base-CoLA],  trained on the CoLA dataset on grammatical acceptability \cite{warstadt2019neural}. Spearman's Rank correlation between the methods and human judgments is reported in Table~\ref{app:fleuncy}. In general, we see the highest correlation with the Cola classifier to human judgment on fluency.

\begin{table}[t] 
\small 
\centering 
\begin{tabular}{@{}lllllllll@{}} 
\toprule \textbf{} & \textbf{} &  \textbf{Lai}  & \textbf{Alva-M.}  & \textbf{Scia.}  & \textbf{Zieg.} & \\ 
\midrule & & spear.  & spear. & spear.  & spear. \\
ppl & & 0.45  & 0.36  & 0.24* & \textbf{0.23} \\ 
clf\_cola & & \textbf{0.52} & \textbf{0.43}& \textbf{0.49}  & 0.20 \\ 
\bottomrule 
\end{tabular} 
\caption{Spearman's Rank correlation between human judgement and a method's predictions. * not significant.} 
\label{app:fleuncy} 
\end{table}

\section{Lincense}
Our test set consists of sentences from Wiki News \url{megarhyme.com/blog/wikinews-dataset/} and from the work of \citet{logacheva-etal-2022-paradetox}. These datasets, respectively, has the licenses Creative Commons Attribution 2.5 and CC0 1.0 Universal. We release our work (both code and dataset) under Creative Commons Attribution 4.0.

\end{document}